# Dual-Class Prompt Generation: Enhancing Indonesian Gender-Based Hate Speech Detection through Data Augmentation

Muhammad Amien Ibrahim
School of Computer Science
Bina Nusantara University
Jakarta, Indonesia
muhammad.amien@binus.ac.id

Faisal
School of Computer Science
Bina Nusantara University
Jakarta, Indonesia
faisal@binus.ac.id

Tora Sangputra Yopie Winarto
School of Computer Science
Bina Nusantara University
Jakarta, Indonesia
tora.winarto@binus.ac.id

Zefanya Delvin Sulistiya
School of Computer Science
Bina Nusantara University
Jakarta, Indonesia
zefanya.sulistiya@binus.ac.id

*Abstract*— Detecting gender-based hate speech in Indonesian social media remains challenging due to limited labeled datasets. While binary hate speech classification has advanced, a more granular category like gender-targeted hate speech is understudied because of class imbalance issues. This paper addresses this gap by comparing three data augmentation techniques for Indonesian gender-based hate speech detection. We evaluate backtranslation, single-class prompt generation (using only hate speech examples), and our proposed dual-class prompt generation (using both hate speech and non-hate speech examples). Experiments show all augmentation methods improve classification performance, with our dual-class approach achieving the best results (88.5% accuracy, 88.1% F1-score using Random Forest). Semantic similarity analysis reveals dual-class prompt generation produces the most novel content, while T-SNE visualizations confirm these samples occupy distinct feature space regions while maintaining class characteristics. Our findings suggest that incorporating examples from both classes helps language models generate more diverse yet representative samples, effectively addressing limited data challenges in specialized hate speech detection.

## I. INTRODUCTION

Online hate speech has emerged as a significant challenge across social media platforms, leading to societal conflicts [1], [2], with gender-based hate speech becoming increasingly prevalent [3], [4]. While platform providers and governments can implement content moderation measures, manual moderation is unfeasible due to high post volumes. Consequently, automated detection systems leveraging machine learning have emerged as practical solutions [5]–[8]. Machine learning approaches for hate speech detection [2], [7], [9] require substantial, balanced training data to prevent classification bias. However, real-world social media data naturally exhibits significant class imbalance, with non-hate speech content considerably outnumbering hate speech posts.

This challenge is particularly evident in Indonesian hate speech research [10], especially in more specific categories like gender-based hate speech. Most studies have focused on binary classification [11]–[13], with limited exploration of specific categories such as gender-based hate speech detection [14], abusive vs hate speech detection [15], and individual vs group hate speech detection [16]. While the dataset in [10] includes gender-based hate speech labels, its severely imbalanced distribution has impeded the development of effective detection systems. Although collecting additional data to balance the classes is possible, such efforts require substantial resources for data collection and annotation. Therefore, computational data augmentation techniques present a promising alternative solution.

Data augmentation techniques include Easy Data Augmentation (EDA) [17] which implements text manipulation operations like synonym replacement and random insertion, demonstrating effectiveness across various domains [18]–[20]. The principles underlying these operations parallel established image augmentation techniques such as flipping, rotation, scaling, and color adjustment, which have become standard practices in computer vision. These EDA methods offer advantages in simplicity and efficiency for low-resource scenarios.

Recently, prompt engineering has emerged as a promising data augmentation approach [21]–[23]. Research by [22] demonstrates the effectiveness of carefully crafted prompts in generating high-quality synthetic data, where carefully structured prompt guide the generation process to maintain consistency and relevance. Similarly, studies have successfully applied this approach for text generation tasks [24], highlighting its potential for hate speech data augmentation. However, previous studies only considered samples from the hate speech class to generate new hate speech examples in a few-shot settings. This study investigates whether using examples from both classes (hate speech and non-hate speech) can improve sample generation for gender-based hate speech detection in Indonesian social media. We contribute to the field by: (1) Comparing our dual-class prompt generation approach with single-class prompting and traditional augmentation techniques, (2) Analyzing the quality and diversity of augmented samples through semantic similarity and distribution visualization, and (3) Evaluating how different augmentation strategies affect classification performance across multiple machine learning models.

## II. LITERATURE REVIEW

Hate speech encompasses harmful communications targeting individuals or groups based on specific characteristics such as race, religion, disability, and gender [15]. Research efforts to identify hate speech on online platforms have been extensive [25]–[27], with growing attention to detection systems for non-English languages [13], [28], [29]. Indonesian hate speech detection research has progressed significantly through a comprehensive dataset [10] containing over 10,000 multi-label samples with various labels. While various machine learning approaches have been explored, dataset quality issues, particularly class imbalances in gender-based hate speech [24] remain challenging.

Data augmentation represents a systematic approach to artificially expanding training datasets while preserving essential characteristics and class distributions. In the context of Indonesian gender-based hate speech detection, previous work has semployed various conventional augmentation techniques, primarily focusing on word-level modifications and document-level modifications [14]. Easy Data Augmentation [17] exemplifies word-level modification approach through operations such as synonym replacement, random insertion, random swap, and random deletion. Backtranslation offers an alternative strategy, operating at the document level by translating text to an intermediate language before converting it back to the original language [30]. This process leverages translation models to introduce linguistic variations while maintaining semantic integrity, a technique successfully applied in hate speech detection to expand training datasets [14].

Recently, prompt engineering has emerged as a state-of-the-art approach to data augmentation, although its application in hate speech detection remains relatively unexplored. While several studies have demonstrated the effectiveness of prompt engineering for data augmentation in various domains [21]–[23], its application to hate speech detection has been limited. A notable exception is [24], which employed prompt engineering to address class imbalance in gender-based hate speech detection. However, opportunities remain for more rigorous evaluation of the quality and effectiveness of samples generated through this approach.

### III. PREPARE YOUR PAPER BEFORE STYLING

This section discusses how the experiment is designed and the methods used with justification for the steps taken in the experiment.

#### A. Dataset

The dataset utilized in this study was obtained from previous research by [31], which collected multi-label hate speech from social media platforms. The dataset comprises 13169 tweets gathered during a period between March to September 2018. Table 1 presents the statistical distribution of the data.

TABLE I. CHARACTERISTICS OF THE DATASET

|  | Non-Gender-Based Hate Speech | Gender-Based Hate Speech |
|---|---|---|
| Total | 12863 | 306 |
| Total (%) | 98 | 2 |

While the dataset contains multiple hate speech categories, this study specifically focuses on gender-related hate speech, examining whether tweets contain gender-targeted hostile content. However, as shown in Table 1, there exists a significant class imbalance, with non-gender-based hate speech tweets 12,863 substantially outnumbering gender-based hate speech tweets 306. It is important to note that 12,863 samples are accumulated from non-hate speech tweets and hate speech tweets that do not fall into gender-based hate speech cateogry. This severe class imbalance poses challenges for model training, potentially leading to biased predictions favoring the majority class and poor detection of gender-based hate speech instances. To address this imbalance, we first transform the dataset into a balanced binary classification task by utilizing all 306 gender-based hate speech samples and randomly selecting an equal number (306) of non-gender-based hate speech samples. This balanced subset serves as our baseline dataset, upon which we apply various data augmentation techniques to expand the minority class while maintaining class balance.

#### B. LLM-based Data Augmentation.

This study explores two prompt-based data augmentation approaches using GPT Large Language Model (LLM). The first approach, which we refer to as single-class prompt generation, utilizes the augmented dataset from [24]. In their work, they generated additional hate speech samples by providing examples of hate speech tweets to the language model, continuing this process until achieving class balance. We use their augmented dataset as one of our comparison points.

Our proposed approach, dual-class prompt generation, introduces a novel perspective by incorporating both hate speech and non-hate speech examples in the prompt which was adopted from [21]. For each generation, we provide the model with five random samples from each class, demonstrating the linguistic characteristics of both hate and non-hate speech. Using GPT-3.5-turbo with carefully controlled parameters (temperature = 0.25, top_p = 0.4) which were chosen based on the best resulting model performances in [24], we generated 306 new samples to match the original class size. The complete prompt template used for generation is provided in the Appendix.

#### C. Backtranslation

The second approach explores conventional data augmentation techniques which is backtranslation. Backtranslation involves translating a tweet into another language and then back to its original language where English was used in this experiment. This process potentially modifies the sentence structure during the translation phases while preserving the semantic meaning. Previous studies have demonstrated significant improvements using this technique [14], [19].

#### D. Preprocessing and Modeling

The dataset underwent standard preprocessing steps, including removal of special characters and normalization of numerical values to "[NUM]" and usernames to "[USERNAME]". We employed traditional machine learning models: Logistic Regression, Naive Bayes, Random Forest, and XGBoost, combined with TF-IDF vectorization. Text preprocessing involved TF-IDF vectorization, which converts tweets into numerical features based on term frequency and inverse document frequency. This approach is particularly used to provide a clear comparison with the works in [24] which used the same approach.

*E. Evaluation*

We used three different ways to check how well our methods worked.

*a) Model Performance Evaluation:* First, we tested how well our models could classify tweets using four different datasets: 1) Original unbalanced data, 2) Data with backtranslation, 3) Data with single-class prompt generation, and 4) Data with our new dual-class prompt generation. We trained four types of classifiers (Logistic Regression, Naive Bayes, Random Forest, and XGBoost) with TF-IDF to convert tweets into numbers. We used 5-fold cross-validation because our dataset is small. Both accuracy and F1-scores were used to measure and evaluate performance. Table 2 summarizes the complete experimental setup of the of the experiment and the evaluation.

TABLE II. SUMMARY OF THE EXPERIMENT SCENARIO

| Dataset | Model | Evaluation |
|---|---|---|
| Original | Logistic Regression Naive Bayes Random Forest XGBoost | Accuracy F-1 Score |
| Original + Backtranslation | | |
| Original + Single-class prompt generation | | |
| Original + Dual-class prompt generation | | |

*b) Semantic Similarity Analysis.* Next, we wanted to see if our generated tweets were actually different from the original ones or just copies with small changes. We used IndoBERTweet [32] which understands Indonesian Twitter language to convert all tweets into number vectors. Then we calculated the average vector for original hate speech tweets and compared it with the average vector for each type of generated tweets using cosine similarity. If the similarity is low, it means the generated tweets contain new patterns and expressions.

*c) Distributional Visualization.* Finally, we created visual maps using T-SNE to see where generated tweets sit compared to original ones. This helps us see if new tweets explore different areas or just cluster with existing tweets. Good generation should create tweets that still look like hate speech but cover areas the original dataset missed. These pictures help us understand patterns that numbers alone might not show.

## IV. RESULTS AND DISCUSSION

### A. Dataset Composition.

Our experiment utilized both original and augmented data as shown in Table 2.

TABLE III. DATASET COMPOSITION

| Source | Non-gender-based HS | Gender-based HS |
|---|---|---|
| Original | 612 | 306 |
| Backtranslation | 0 | 306 |
| Single-class prompt generation | 0 | 306 |
| Dual-class prompt generation | 0 | 306 |

Each augmentation method produced 306 new gender-based hate speech samples, which when combined with original samples created balanced datasets for training.

### B. Classification Performance

We evaluated model performance across different dataset configurations using accuracy and F1-score metrics with 5-fold cross-validation. Results are presented in Table 4. The results show that all augmentation methods improve model performance compared to training on original data alone. Our dual-class prompt generation method achieved the highest performance with Random Forest (accuracy 0.885, F1-score 0.881), XGBoost (accuracy 0.882, F1-score 0.887), and Logistic Regression (accuracy 0.871, F1-score 0.861). The single-class prompt generation method produced the second-best results, while backtranslation showed the smallest improvement. A notable observation is the gap between accuracy and F1-score in the original dataset (e.g., 0.798 vs 0.590 for Logistic Regression), which narrows significantly with augmented data. This suggests that augmentation helps models better identify both classes rather than favoring the majority class. In terms of Naive Bayes, this model performed best with backtranslation (accuracy 0.847, F1-score 0.856), indicating this model may benefit from the linguistic variations introduced by translation.

### C. Semantic Similarity Analysis

To assess whether our augmentation methods generate novel content, we calculated cosine similarity between original hate speech samples and augmented samples using IndoBERTweet embeddings:

TABLE IV. SEMANTIC SIMILARITY SCORES

| Augmentation Method | Similarity to Original |
|---|---|
| Backtranslation | 0.9303 |
| Single-class prompt generation | 0.9635 |
| Dual-class prompt generation | 0.8684 |

Lower similarity scores indicate more novel content. Our dual-class prompt generation technique produced the most semantically different content (0.8684), while single-class prompt generation created content most similar to the original (0.9635). This suggests that incorporating non-hate speech examples in the prompt helps the model generate more diverse hate speech samples.

### D. Distribution Visualization

T-SNE visualizations revealed different distribution patterns for each augmentation method as shown in Figures 1-4.

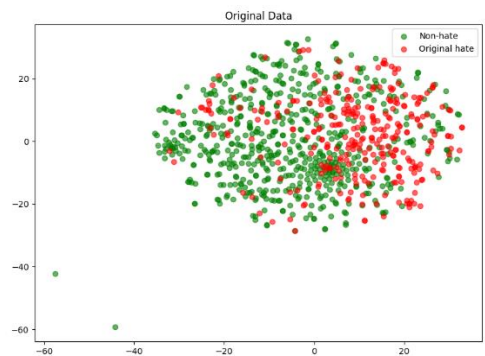

Figure 1. T-SNE visualization of original dataset

TABLE V. MODEL PERFORMANCE ACROSS DATASET CONFIGURATIONS

| Dataset | Model | Accuracy | Accuracy Std | F1-Score | F1-Score Std |
|---|---|---|---|---|---|
| Original | Logistic Regression | 0.798 | 0.025 | 0.590 | 0.068 |
| | Naive Bayes | 0.777 | 0.020 | 0.518 | 0.060 |
| | Random Forest | 0.849 | 0.027 | 0.729 | 0.054 |
| | XGBoost | 0.843 | 0.023 | 0.736 | 0.043 |
| Backtranslated | Logistic Regression | 0.829 | 0.027 | 0.827 | 0.026 |
| | Naive Bayes | **0.847** | 0.029 | **0.856** | 0.024 |
| | Random Forest | 0.820 | 0.037 | 0.817 | 0.034 |
| | XGBoost | 0.797 | 0.032 | 0.790 | 0.033 |
| Single-class prompt generation | Logistic Regression | 0.858 | 0.018 | 0.849 | 0.020 |
| | Naive Bayes | 0.835 | 0.027 | 0.849 | 0.020 |
| | Random Forest | 0.884 | 0.021 | 0.877 | 0.023 |
| | XGBoost | 0.877 | 0.017 | 0.870 | 0.020 |
| Dual-class prompt generation | Logistic Regression | **0.871** | 0.023 | **0.861** | 0.028 |
| | Naive Bayes | 0.833 | 0.027 | 0.844 | 0.020 |
| | Random Forest | **0.885** | 0.019 | **0.881** | 0.019 |
| | XGBoost | **0.882** | 0.020 | **0.877** | 0.022 |

Figure 1 displays the original dataset where gender-based hate speech samples (red) appear scattered across the feature space with some overlapping with non-gender hate speech (green). This overlap indicates potential labeling challenges or similar linguistic patterns between some samples of different classes.

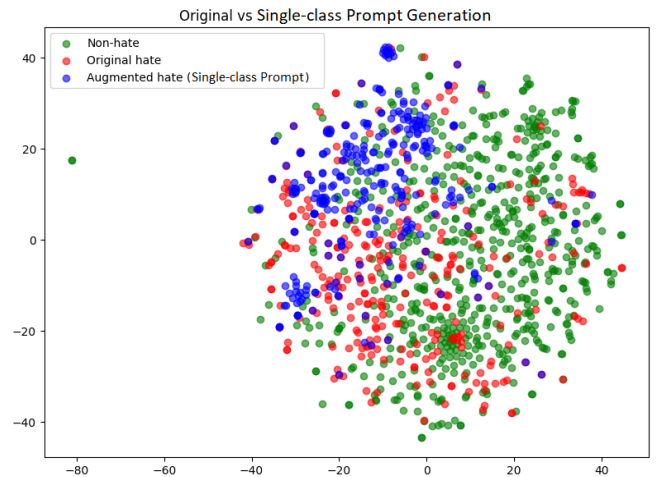

Figure 3. T-SNE visualization of original data vs single-class prompt generation

In Figure 3, we observe the single-class prompt generated samples compared to the original dataset. These generated samples form more distinct clusters while still maintaining some overlap with original samples. The clustering pattern suggests that this generation method tends to produce samples with more consistent linguistic patterns than backtranslation.

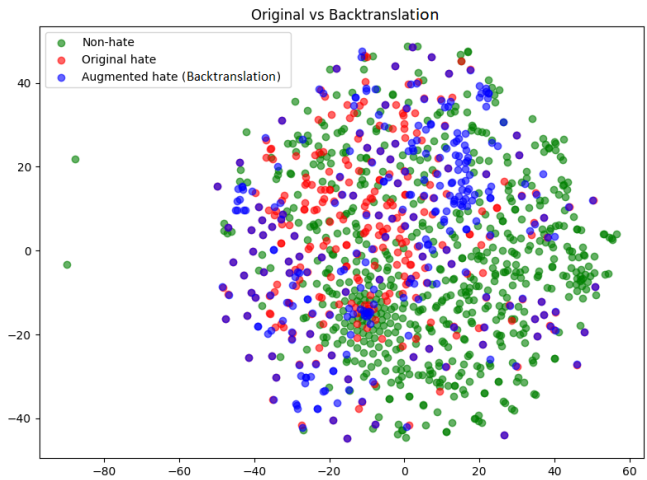

Figure 2. T-SNE visualization of original data vs Backtranslation

Figure 2 shows the distribution of backtranslated samples (blue) compared to the original data. The backtranslated samples appear scattered throughout the feature space with partial overlap with original samples. This distribution demonstrates backtranslation's ability to generate novel content while maintaining some characteristics of the original samples.

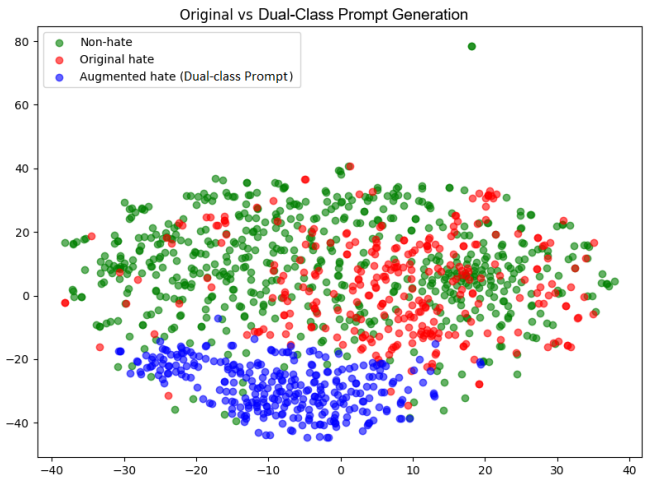

Figure 4. T-SNE visualization of original data vs dual-class prompt generation

Figure 4 presents the dual-class prompt generated samples alongside the original data. These samples demonstrated the highest degree of clustering and least overlap with original samples. This distinct positioning in the feature space suggests that our dual-class prompt generation approach produces the most novel content while still maintaining the essential characteristics of gender-based hate speech, supporting our findings from the semantic similarity analysis.

## V. CONCLUSION

Our study examined three data augmentation techniques for Indonesian gender-based hate speech detection: backtranslation, single-class prompt generation, and our proposed dual-class prompt generation. Through comprehensive evaluation using model performance metrics, semantic similarity analysis, and distribution visualization, we found that all augmentation methods improved classification performance compared to using only original data.

The dual-class prompt generation technique demonstrated the best overall performance, achieving the highest accuracy and F1-scores with multiple classifiers. This approach also produced the most semantically novel content as measured by cosine similarity (0.8684) and created samples that occupied distinct regions in the feature space. Our findings suggest that providing both positive and negative examples to guide generation helps create more diverse yet representative samples.

While all three augmentation techniques proved beneficial, they showed different strengths. Backtranslation performed particularly well with Naive Bayes classifiers, suggesting it introduces linguistic variations that benefit certain modeling approaches. Single-class prompt generation showed strong performance but produced samples most similar to the original data.

The limitations of this study include the relatively small dataset size and focus on a single language and platform (Indonesian Twitter). Future work should explore these augmentation techniques on larger datasets, different languages, and additional hate speech categories. Additionally, investigating how these methods perform when combined could reveal potential complementary benefits.

Our research contributes to addressing the challenge of limited labeled data in hate speech detection, particularly for low-resource languages and specific hate speech categories. The proposed dual-class prompt generation technique offers a promising approach for creating more balanced datasets while maintaining linguistic diversity.

## VI. FUNDING INFORMATION

This study is supported by Research and Technology Transfer Office, Bina Nusantara University as a part of Bina Nusantara University's Associate Research Grant entitled Prompt-Based Data Augmentation with Large Language Models for Indonesian Gender-Based Hate Speech Detection with contract number: 069A/VRRTT/III/2024 and contract date: March 18, 2024.

APPENDIX

```
The following tweets belong to the category of 'non-hate speech towards gender':
1. {example1}
2. {example2}
3. {example3}
4. {example4}
5. {example5}

The following tweets belong to the category of 'hate speech towards gender':
1. {example1}
2. {example2}
3. {example3}
4. {example4}
5. {example5}

Please generate a new tweet that belongs to the category of 'hate speech towards gender'. Important requirement: Generate the tweet in Indonesian language.
Generated tweet:
```

Figure 5. The complete prompt template used for generation